%% file: root.tex
\documentclass[letterpaper, 10 pt, journal, twoside]{IEEEtran}

\IEEEoverridecommandlockouts

\input{meta/packages}

\input{meta/usercmds}

\usetikzlibrary{shapes.geometric, arrows.meta, positioning}

\title{CLEAR: A Semantic–Geometric Terrain Abstraction for Large-Scale Unstructured Environments}

\author{Pranay Meshram$^{1}$, Charuvahan Adhivarahan$^{1}$,
Ehsan Tarkesh Esfahani$^{2}$, Souma Chowdhury$^{2}$, \\
Chen Wang$^{1}$, and Karthik Dantu$^{1}$%
\thanks{Manuscript received: December 29, 2025; Revised: May 26, 2026; Accepted: August 6, 2026.}%
\thanks{This paper was recommended for publication by Editor Giuseppe Loianno upon evaluation of the Associate Editor and Reviewers' comments. This work was partially supported by the Office of Naval Research (ONR) under Grant No. N00014-24-1-2003 and a gift from MOOG Inc.}%
\thanks{$^{1}$Department of Computer Science and Engineering, University at Buffalo, NY 14260, USA.
{\tt\footnotesize \{pmeshram, charuvah, cwx, kdantu\}@buffalo.edu}}%
\thanks{$^{2}$Department of Mechanical and Aerospace Engineering, University at Buffalo, NY 14260, USA.
{\tt\footnotesize \{ehsanesf, soumacho\}@buffalo.edu}}%
\thanks{Digital Object Identifier (DOI): see top of this page.}%
}

\begin{document}

\maketitle

\input{sec/abstract}

\input{sec/intro}

\input{sec/related}

\input{sec/problem_formulation}

\input{sec/Methodology}

\input{sec/evaluation_and_results}

\input{sec/conclusion}

\bibliographystyle{IEEEtran}
\bibliography{references}

\end{document}

%% file: meta/packages.tex
\usepackage[T1]{fontenc}
\usepackage{soul}
\setstcolor{green}
\usepackage{graphicx} %

\usepackage[caption=false,font=footnotesize]{subfig}

\usepackage{algorithm}

\usepackage{algpseudocode}
\usepackage{amsmath}
\usepackage{mathtools}
\usepackage{hyperref}

\usepackage{amssymb}       %
\usepackage{pifont}        %

\newcommand{\cmark}{\ding{51}}  %
\newcommand{\xmark}{\ding{55}}  %

\usepackage[table]{xcolor}
\usepackage{tikz}

\usepackage{siunitx}
\usepackage[acronym]{glossaries-extra}

\usepackage{booktabs}
\usepackage{multirow}

\usepackage{cite}
\usepackage{enumitem}

%% file: meta/usercmds.tex
\newcommand{\revise}[1]{#1}

\newlist{enumlite}{enumerate}{3}
\setlist[enumlite]{label=\arabic*),wide,labelindent=0pt}

\newlist{enumline}{enumerate*}{3}
\setlist[enumline]{label=(\arabic*), itemjoin*={{, and }}}

\newlist{itemlite}{itemize}{3}
\setlist[itemlite]{label=\textbullet,wide,labelindent=0pt,leftmargin=*}

\glsdisablehyper
\newabbreviation{clear}{CLEAR}{Connected Landcover Elevation Abstract Representation}

%% file: sec/abstract.tex
\begin{abstract}

Long-horizon navigation in unstructured environments demands terrain abstractions that scale to tens of square kilometers while preserving semantic and geometric structure. Grids scale poorly and quadtrees misalign with terrain boundaries. Neither encodes terrain semantics essential for traversability-aware planning, yielding infeasible or inefficient paths for autonomous ground vehicles operating over 10+~km$^2$.
\textbf{CLEAR} (Connected Landcover Elevation Abstract Representation) is a reusable terrain abstraction framework for large-scale planning producing convex, semantically aligned regions encoded as a terrain-aware graph.
Evaluated on 9--100~km$^2$ digital terrain maps with physics-based simulation, CLEAR reduces per-query planning time by \textbf{$2\times$ to $19.6\times$ over unabstracted raw-grid A*} after one-time abstraction with \textbf{6.7\% cost overhead}. In physics-based simulation, CLEAR delivers \textbf{5--8.6\% shorter executed paths and 100\% task completion across all maps versus 90--100\% for Quadtree}. These results hold against AMRA*, a strong anytime multi-resolution planner, and generalize to a learned elevation-driven cost, demonstrating CLEAR's scalability and cost-model independence as a reusable planning layer.
We release our code and dataset at \url{https://github.com/droneslab/CLEAR}.
\end{abstract}


%% file: sec/intro.tex
\section{Introduction}

\IEEEPARstart{A}{utonomous} navigation in unstructured environments requires reasoning on large, high-resolution maps of landcover and elevation, which is computationally prohibitive at scale.
Unlike on-road navigation, where predefined road networks provide a strong planning graph, off-road environments lack such structure and require abstractions that infer traversable connectivity from terrain itself.
Traditional abstractions such as grid, hex, or quadtree discretizations simplify maps but often ignore semantic terrain boundaries and geometric structure, leading to inefficient planning or infeasible paths.

\revise{Consider autonomous wildfire response over 50~km$^2$ of mixed terrain. Grids are intractable at this scale, and quadtrees ignore landcover, yielding routes that violate slope or friction constraints. Recent efforts in wildfire robotics ~\cite{seraj2022multi, al2024hierarchical, 10876021, jindal2021design} highlight the need for scalable terrain abstractions, yet prior work focuses primarily on per-query planning. In this work, CLEAR introduces a reusable planning abstraction layer.}
\revise{We target ground vehicles, whose mobility is directly constrained by slope, friction, and landcover---factors less central to aerial systems.}

\begin{figure}[ht]
    \centering

    \includegraphics[width=\linewidth]{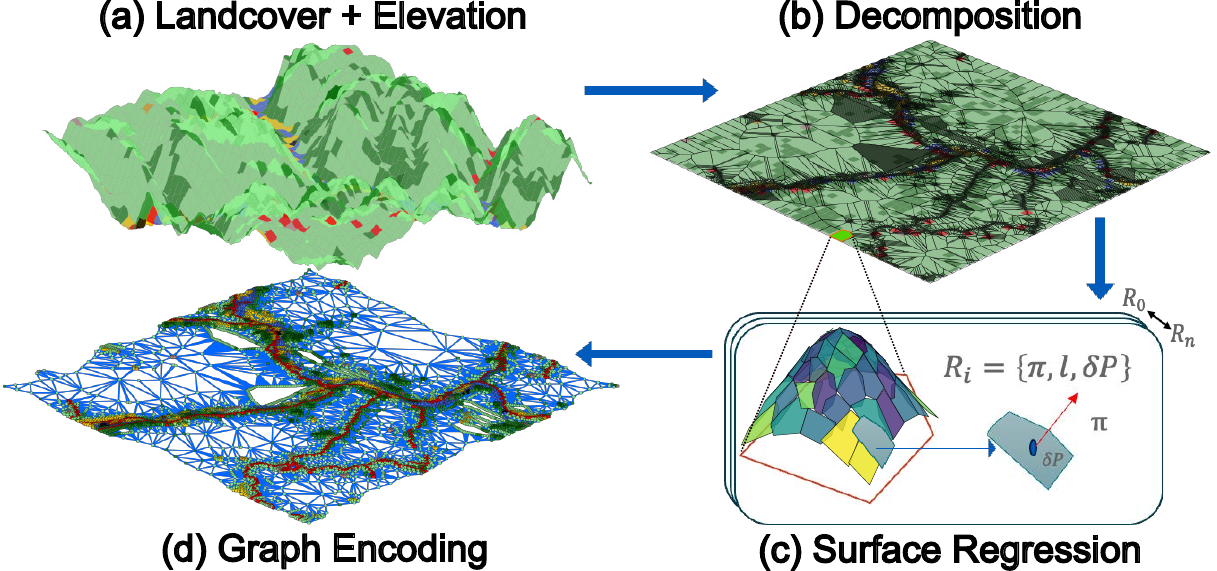}

    \vspace{-1mm}
    \caption{\revise{CLEAR pipeline: (a) landcover and elevation, (b) Boundary-Seed Decomposition, (c) recursive planar fitting, and (d) terrain-aware graph encoding.}}
    \label{fig:clear_summary}

\end{figure}

\revise{While \gls{clear} uses known primitives, it solves a distinct problem: scalable terrain abstraction for reusable planning. After one-time preprocessing, repeated queries search an $O(\text{graph size})$ region graph rather than the $O(\text{map size})$ raw raster.}
\revise{\gls{clear} advances reusable terrain planning abstraction through:}
\begin{enumlite}
    \item \revise{\textbf{Reusable and repeatable abstraction:} Builds a pre-mission terrain graph reused across queries, with repeatable regions supporting map stitching and multi-robot planning.}
    \item \revise{\textbf{Planning-aware convex decomposition:} Jointly optimizes for landcover consistency, elevation geometry, convex regions, and terrain-cost alignment.}%
    \item \revise{\textbf{Multi-scale amortized validation:} Evaluated on 9--100~$\mathrm{km}^2$ terrains with amortized planning, physics-based simulation, and Jetson AGX Orin timing.}

\end{enumlite}

%% file: sec/related.tex
\section{Related Work}
\label{sec:related_work}

\textbf{Terrain Representation and Abstraction}:
Classical abstractions such as grids ~\cite{hart1968formal, ball2016vision}, hexagonal tilings~\cite{duszak2021hexagonal}, and quadtrees~\cite{hart1968formal} simplify large maps but remain misaligned with semantic terrain boundaries.
\revise{Adaptive image-segmentation methods such as SLIC superpixels~\cite{achanta2012slic} provide a stronger boundary-aware decomposition reference than uniform grids, but they cluster pixels by spatial proximity and color similarity and do not explicitly optimize for terrain semantics, elevation-derived geometry, or graph-planning attributes.}
In contrast, indoor semantic scene graphs---e.g., 3D Scene Graph~\cite{strader2024indoor}, ConceptGraphs~\cite{10610243}, and Hydra/Multi-Hydra~\cite{hughes2022hydra,chang2023hydra}---enable rich reasoning but operate at building scale with object- or room-level semantics.
CLEAR develops an analogous abstraction to outdoor terrains by coupling semantic landcover boundaries with elevation-derived geometry---a combination absent in prior work.

\textbf{Surface Simplification and Plane Fitting}:
Surface-fitting models such as curved or planar patches~\cite{kanoulas2019curved}, seed-curve fitting in agriculture~\cite{jin2011coverage}, and adaptive quadtree surfaces~\cite{zhang2025optimization} compactly approximate terrain geometry but ignore semantic consistency.
CLEAR extends these ideas by applying recursive plane fitting within semantically coherent regions, preserving both geometric fidelity and semantic alignment.

\textbf{Path Planning on Abstract Maps}:
Standard planners (A*~\cite{hart1968formal}, RRT*~\cite{karaman2011sampling}) ignore terrain-specific costs like slope and friction.
\revise{HPPRM~\cite{ravankar2020hpprm} improves roadmap planning with hybrid potential fields but is formulated around 2D obstacle proximity, not 2.5D terrain costs (elevation, grade, landcover).
Geometry-aware frameworks such as PUTN~\cite{jian2022putn} and continuous surface maps~\cite{zhang2025optimization} address uneven terrain geometry but are not drop-in graph-search planners over an abstract region graph.
Learning-based terrain planners such as TERP~\cite{weerakoon_terp_2022} learn elevation-driven cost priors for uneven outdoor navigation and are complementary to \gls{clear}, since \gls{clear} abstracts the map structure and can evaluate paths under such learned cost maps.
Anytime and multi-resolution methods such as ARA*~\cite{likhachev2003ara} and AMRA*~\cite{saxena2022amra} target grid-search acceleration via cell aggregation, not semantic terrain abstraction. Their nodes remain terrain-blind at every resolution level.
}

\revise{Thus, prior work offers geometry-only terrain models, appearance-driven segmentations, grid-based search accelerations, or per-query planners. \gls{clear} targets the missing abstraction layer: a reusable, semantic--geometric, planning-aware region graph for large-scale off-road terrain.}

%% file: sec/problem_formulation.tex
\glsreset{clear}
\section{\texorpdfstring{\gls{clear}}{CLEAR}}

\subsection{Problem Formulation}

Let \( \Omega \subset \mathbb{R}^2 \) denote the terrain. For each point \(p = (x, y) \in \Omega\):
\begin{itemlite}
    \item \(E(p) \in \mathbb{R}\) gives the elevation,
    \item \(L(p) \in C\) gives the landcover class, where \(C = \{c_1,\dots,c_N\}\).
\end{itemlite}

\revise{CLEAR decomposes \(\Omega\) into a collection of convex, semantically coherent regions \(\{R_k\}_{k=1}^K\) produced by the Boundary-Seed Decomposition (BSD) described in Sec.~\ref{sec:bsd}.}
Each region is then processed by recursive plane fitting, which may further subdivide \(R_k\) into planar subsets.
Each resulting planar region is represented by:
\[
R_k = (\pi_k, \mathcal{P}_k, c_k),
\]
where \( \pi_k \) is the tangent plane fitted to the elevation values in that planar subset,
\( \mathcal{P}_k \) is its polygonal boundary, coinciding with the original BSD polygon only when no subdivision is required,
and \( c_k \in C \) is the dominant landcover label of the points belonging to that region.

\noindent Each region satisfies:
\begin{enumlite}
    \item \textit{Geometric fidelity:} the RMSE between \(E\) and \(\pi_k\) over the region is below a tolerance \(\epsilon\).
    \item \textit{Semantic uniformity:} \(c_k\) is the majority landcover label among the points in the region.
    \item \revise{\textit{Structural convexity:} BSD ensures convex regions, and recursive splitting preserves convex subregions with boundary \(\mathcal{P}_k\). This supports graph search over region adjacency and leaves a direct path to continuous optimization methods such as Graphs of Convex Sets~\cite{marcucci2024shortest}.}
\end{enumlite}

The final abstraction defines a region graph \(\mathcal{G} = (\mathcal{V}, \mathcal{E})\), where vertices represent regions and edges encode traversable adjacency.

%% file: sec/Methodology.tex
\begin{table}[ht]
\centering
\footnotesize
\renewcommand{\arraystretch}{1.1}
\caption{\revise{Spatial decomposition attributes for reusable terrain planning. SLIC denotes superpixel~\cite{achanta2012slic} and BSD denotes the proposed Boundary-Seed Decomposition.}} %
\label{tab:decomposition_comparison}
\setlength{\tabcolsep}{6pt}
\resizebox{\linewidth}{!}{%
\begin{tabular}{|l|c|c|c|c|c|}
\hline
\textbf{Attribute} & \textbf{Grid} & \textbf{Hex} & \textbf{SLIC} & \textbf{Quadtree} & \textbf{BSD} \\
\hline
\textbf{[A1]} Convex region support     & \cmark & \cmark & \xmark & \cmark & \cmark \\
\textbf{[A2]} Planner compatibility     & \cmark & \cmark & \xmark & \cmark & \cmark \\
\textbf{[A3]} Adaptive resolution       & \xmark & \xmark & \xmark & \cmark & \cmark \\
\textbf{[A4]} Compact representation    & \xmark & \xmark & \cmark & \cmark & \cmark \\
\textbf{[A5]} Semantic alignment        & \xmark & \xmark & \xmark & \xmark & \cmark \\
\textbf{[A6]} Geometry-aware fitting    & \xmark & \xmark & \xmark & \xmark & \cmark \\
\textbf{[A7]} Interpretability          & \xmark & \xmark & \xmark & \xmark & \cmark \\
\textbf{[A8]} Repeatability             & \xmark & \xmark & \xmark & \xmark & \cmark \\
\hline
\end{tabular}
}
\end{table}

\revise{Table~\ref{tab:decomposition_comparison} compares spatial decompositions by the properties needed for reusable terrain planning. Interpretability~(A7) means that each region exposes an inspectable boundary, dominant landcover label, and elevation-derived attributes used by the graph cost. BSD is the only decomposition that satisfies all eight attributes, making it the foundation of \gls{clear}'s planning-aware region graph.}

\subsection{\texorpdfstring{\gls{clear}}{CLEAR} Design}

CLEAR operates through:
\begin{enumline}
    \item \textit{BSD}: region formation aligned with terrain boundaries,
    \item \textit{Plane Fitting}: geometric simplification within each region
    \item \textit{Graph Encoding}: building a region-level adjacency structure for planning tasks.
\end{enumline}

\begin{figure}[!t]
    \centering
    \includegraphics[width=\linewidth]{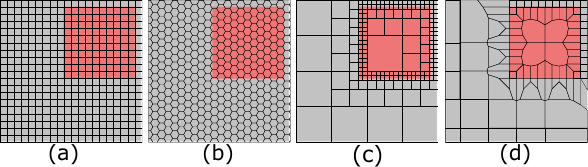}
\vspace{-6mm}
\caption{\revise{Example decompositions of the same terrain: (a) Grid, 400 cells, (b) Hexagonal, 448 cells, (c) Quadtree, 184 cells, and (d) BSD, 105 cells.}}
\label{fig:region-decompositions}
\end{figure}

\subsubsection{Boundary-Seed Decomposition}
\label{sec:bsd}
\begin{algorithm}
\caption{Boundary-Seed Decomposition}
\label{alg:bas}
\begin{algorithmic}[1]
\Require
$L\in\mathbb{Z}^{H\times W}$ (landcover map),\;
$E\in\mathbb{R}^{H\times W}$ (elevation map),\;
target $n$ (total seeds),\;
$\alpha_{\text{bdy}}\in[0,1]$ (boundary priority),\;
$r_{\min}$ (minimum inter-seed distance),\;
$k$ (local entropy window size)
\Ensure $\mathcal{S}$ with $|\mathcal{S}|=n$
\State $\sigma_E \gets \textsc{LocalStd}_k(E)$;\quad $\tau_{\text{flat}} \gets \mathrm{quantile}_{0.25}(\sigma_E)$
\State $B \gets \textsc{BoundaryMask}(L)$;\quad $H_k(p)\gets \textsc{LocalEntropy}_k(L)$;\quad $R \gets \operatorname*{argsort}_{p\in B}^{\downarrow} H_k(p)$
\State $n_f \gets n - (\alpha_{\text{bdy}}\, n)$;\quad $\mathcal{S}\gets\varnothing$
\State \textbf{Uniform flat sampling:} visit pixels by ascending $\sigma_E$; accept $p$ if $\min_{q\in\mathcal{S}}\|p-q\|_2 \ge r_{\min}$ until $|\mathcal{S}|=n_f$
\State \textbf{Flat-region centroids:} for each label $c$, form connected components of $\{\sigma_E \le \tau_{\text{flat}}, L=c\}$, add their centroids until $|\mathcal{S}|\ge n_f$
\State $m \gets n-|\mathcal{S}|$;\quad append the first $m$ elements from $R\setminus\mathcal{S}$ to $\mathcal{S}$
\While{$|\mathcal{S}|<n$}
  \State append next elements from $R\setminus\mathcal{S}$ to $\mathcal{S}$
\EndWhile
\State \Return first $n$ points of $\mathcal{S}$
\end{algorithmic}
\end{algorithm}
\revise{Boundary-Seed Decomposition (BSD) places seeds in flat regions and along high-entropy semantic boundaries, controlled by a priority factor $\alpha_{\text{bdy}}$ as described in Algorithm~\ref{alg:bas}.
\textbf{Key idea:} Classical Voronoi tessellation partitions by distance, while BSD first places seeds by boundary entropy and flatness before partitioning.} \revise{The target seed count \(n\) directly sets the region budget and can help alleviate computational load as validated in evaluation.} Figure~\ref{fig:region-decompositions} shows BSD yields fewer, more coherent partitions than grid, hexagonal, or quadtree methods.

\subsubsection{Elevation Plane Fitting}

To capture terrain geometry, CLEAR approximates elevation within each BSD region using recursive planar fitting.
A plane $\pi_k$ is first fitted to the elevation values of the region. If the RMSE exceeds a tolerance ($\epsilon$), the region is subdivided into four convex subregions and refitted.
Each resulting planar patch inherits its dominant landcover label and is assigned a convex polygonal boundary obtained as the convex hull of its inlier points.
This produces planar regions satisfying tolerance $\epsilon$.
The full procedure is summarized in Algorithm~\ref{alg:plane_fitting}.

\textbf{Complexity.} BSD runs in $O(N \log N)$ for Voronoi construction. Plane fitting terminates when RMSE $\leq \varepsilon$ or area $< A_{\min} = 4$ pixels$^2$, ensuring depth $d \leq \lceil\log_4(A_0/A_{\min})\rceil = O(\log N)$, yielding $O(N \log N)$ overall.

\begin{algorithm}
\caption{Recursive Planar Region Fitting with Landcover Labels}
\label{alg:plane_fitting}
\begin{algorithmic}[1]
\Require
3D points $\mathcal{X} = \{(x_i, y_i, z_i)\}_{i=1}^N$,
landcover labels $\mathcal{L} = \{\ell_i\}_{i=1}^N$,
RMSE threshold $\epsilon$
\Ensure
Region set $\mathcal{R} = \{(\pi_k, \mathcal{P}_k, c_k)\}$,
where $\pi_k$ is a fitted plane,
$\mathcal{P}_k$ is a convex polygon (boundary of the planar patch),
and $c_k$ is the dominant landcover label

\State Initialize queue with $(\mathcal{X}, \mathcal{L})$
\While{queue is not empty}
    \State Pop $(\mathcal{X}_i, \mathcal{L}_i)$ from queue
    \State Fit plane $\pi_i$ to points $\mathcal{X}_i$ via least squares
    \State Compute RMSE: $e_i = \texttt{RMSE}(Z, \pi_i)$

    \If{$e_i \le \epsilon$ or $\mathcal{X}_i$ is spatially small}
        \State $\mathcal{P}_i \gets \texttt{ConvexHull}(\mathcal{X}_i)$ // Clipped to parent
        \State $c_i \gets \texttt{mode}(\mathcal{L}_i)$
        \State Append $(\pi_i, \mathcal{P}_i, c_i)$ to $\mathcal{R}$
    \Else
        \State Compute medians $(x_m, y_m)$ of $\mathcal{X}_i$
        \State Partition $\mathcal{X}_i$ into quadrants
               $\{\mathcal{X}_{i,k}\}_{k=1}^4$ using $(x_m, y_m)$
        \State Push each $(\mathcal{X}_{i,k}, \mathcal{L}_{i,k})$ to queue
    \EndIf
\EndWhile

\State \Return $\mathcal{R}$
\end{algorithmic}
\end{algorithm}

\subsubsection{Graph Encoding}

Planar regions form a graph $G = (V, E)$ where edges $(v_i \rightarrow v_j)$ receive terrain-aware costs from Sec.~\ref{sec:cost_function},
incorporating slope, roughness, landcover, and heading alignment.

\subsection{Map Encoding for Terrain-Aware Path Planning}

Each planar region is annotated with terrain attributes and a traversal cost to enable terrain-aware navigation.

\subsubsection{Encoded Attributes}
\label{subsec:encoded_attributes}
\setlength{\itemindent}{0em}
\begin{itemize}
    \item \textit{Percent Grade:} encodes the magnitude of terrain slope. Raster-step transitions exceeding 35\% grade receive the finite barrier cost used in the planner.
    \item \textit{Slope Direction (Aspect):} Orientation of steepest descent, used for slope-aligned and energy-efficient traversal.
    \item \textit{Elevation Statistics:} Max/mean/min elevation values capturing local relief and terrain variability.
    \item \textit{Landcover Type:} \revise{A categorical raster label \(L\). While experiments use NALCMS, any categorical raster (classifier output, trail map, or mission mask) can be substituted, with coefficients~\cite{pandolf1977,papadakis2013terrain, fu2025anynav} used to favor or block specific classes.}
\end{itemize}
\revise{For regions containing multiple landcover classes, CLEAR assigns the dominant landcover label for graph-cost evaluation. This keeps the region graph compact and interpretable, but can under-represent minority classes inside mixed regions. Boundary-seeded decomposition mitigates this by placing seeds along landcover transitions to reduce label mixing. Remaining mixed-region effects are reflected in the reconstruction mIoU and JSD metrics.}

\subsubsection{Terrain-Aware Cost for Vehicle Path Planning}
\label{sec:cost_function}
\revise{The region graph is designed to accept any pluggable cost function. We validate this with two complementary cost models: the analytic multi-attribute traversal cost of~\cite{fu2025anynav} and TERP's~\cite{weerakoon_terp_2022} elevation-driven learned cost. The former jointly incorporates elevation-derived geometry (slope, roughness) and surface-class properties (friction from landcover), matching CLEAR's dual-attribute regions. We omit its steering-penalty term to avoid over-constraining long-range paths. For TERP, we process the raw elevation map through TERP's learned model to obtain a dense cost surface, which is then aggregated over each abstract region to compute graph edge weights.} We add a finite grade barrier ($B=10^6$ for path segments whose percent grade exceeds 35\%) and an explicit slope--alignment penalty
\begin{equation*}
\Delta \theta =
\min\!\left(
|\theta_{\text{head}} - \theta_{\text{slope}}|,\;
360 - |\theta_{\text{head}} - \theta_{\text{slope}}|
\right),
\label{eq:transfer_cost}
\end{equation*}
\revise{ which penalizes heading mismatch with the local slope.}
Edge cost follows~\cite{fu2025anynav}: $C_{ij} = d_{ij}(1 + w_f\mu_{ij} + w_s|s_j| + w_r\sigma_j)$ with weights from~\cite{fu2025anynav}, plus our heading penalty $w_\theta\Delta\theta/180^\circ$ with $w_\theta=0.1$.

\revise{Results for both cost models are reported in Sec.~\ref{sec:evaluation}.}

%% file: sec/evaluation_and_results.tex
\section{Evaluation and Results}
\label{sec:evaluation}

We evaluate \gls{clear} across three dimensions:
(1)~\textbf{Structural and semantic performance} evaluating geometric fidelity, semantic consistency, %
and repeatability of our terrain abstraction,
(2)~\textbf{Path planning and deployability} benchmarking computational efficiency and path feasibility in simulation, and
(3)~\textbf{Ablation studies} analyzing the sensitivity of decomposition parameters such as region count and boundary emphasis.

\revise{
To the best of our knowledge, \gls{clear} is the first reusable terrain abstraction that jointly encodes landcover semantics and elevation geometry for large-scale off-road planning, making it challenging to find similar baselines to compare with. Therefore, we take a multi-pronged approach. For decomposition quality, we compare against Grid, Hexagonal, Quadtree, and SLIC~\cite{achanta2012slic}. SLIC is a boundary-aware reference sharing a similar region-growing principle as \gls{clear} in image space. For planning, we use raw-raster A* as the optimal reference and RRT* as a sampling-based alternative. PUTN~\cite{jian2022putn} is excluded as it is a full navigation system with its own planning representation and cost interface, whose inclusion would alter the graph-search problem rather than isolate decomposition quality.
For scalability, we compare \gls{clear} with AMRA*~\cite{saxena2022amra} to demonstrate its accuracy and efficiency in planning.
To demonstrate cost-model independence, we substitute the AnyNav-adapted~\cite{fu2025anynav} traversal cost with TERP's~\cite{weerakoon_terp_2022} learned elevation cost. All planning evaluations are conducted on three maps spanning 9--100~km\textsuperscript{2}.}

\textbf{Hyperparameters:}
\revise{In \autoref{alg:bas} (Boundary-Seed Decomposition), two key parameters control seed placement.}
\revise{First, the \emph{boundary emphasis} parameter~\(\alpha_{bdy}\) balances flatness-driven and boundary-driven seeding. We treat it as a task-dependent tradeoff. Decomposition metrics use \(\alpha_{bdy}=1\), planning results use \(\alpha_{bdy}=0\), and Sec.~\ref{subsec:alpha_ablation} reports the resulting sensitivity.}
Second, the \emph{flatness threshold}~\(\tau_{\text{flat}}\) defines locally planar areas eligible for seed selection.
We compute it as \(\tau_{\text{flat}} = P_{25}\!\big(\textsc{LocalStd}_{5\times5}(E)\big)\), corresponding to the flattest quartile (in meters).
We set $k=3$ and $r_{\min}=2\text{--}4$\,px to minimize boundary mixing while maintaining uniform seed density. \(r_{\min}\) scales with pixel size, while \(k\) may be increased to~5 for stronger boundary discrimination.

In \autoref{alg:plane_fitting} (Recursive Plane Fitting), the primary control parameter is the \emph{planarity tolerance}~\(\varepsilon\), set to \(10\,\mathrm{m}\) for \(30\,\mathrm{m}\)~DEMs.
This threshold governs the termination of plane refinement and directly affects the trade-off between geometric accuracy and region compactness in the final abstraction.

All experiments were conducted on a machine running Ubuntu 20.04 LTS with a 12th Gen Intel Core i9-12900K CPU and 32 GB RAM.

\begin{figure*}[!t]

    \centering
    \includegraphics[width=\textwidth]{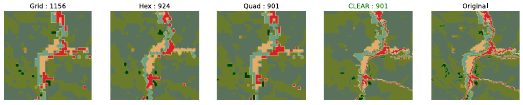}
    \vspace{-7mm}
    \caption{Land-cover reconstruction on the Wharton (W) subregion.
Grid, Hex, and Quadtree distort semantic boundaries, whereas \gls{clear} better preserves fine classes and irregular transitions.}

    \label{fig:landcover-reconstruction}
\end{figure*}

\begin{figure}[!t]
    \centering
    \includegraphics[width=\linewidth]{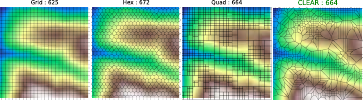}
    \vspace{-6mm}
    \caption{Elevation reconstruction on the Wharton (W) subregion with $\sim$650 regions.
\gls{clear} aligns with terrain contours and yields smooth planar regions, while Grid/Hex show stair-step artifacts and Quadtree exhibits block artifacts.}

    \label{fig:elevation-reconstruction}
\end{figure}

\begin{figure}
    \centering
    \includegraphics[width=\linewidth] {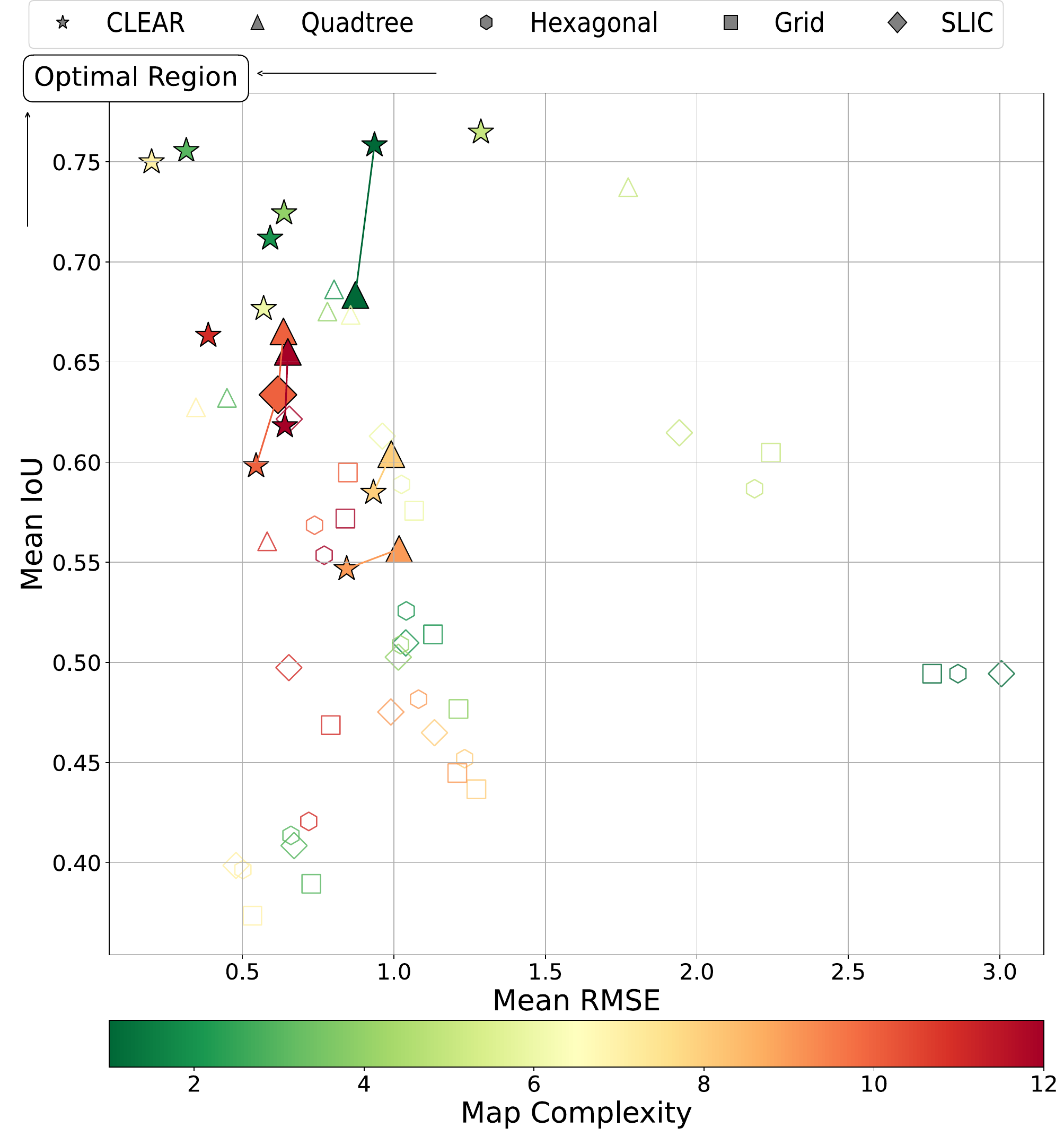}
    \vspace{-2mm}
    \caption{\revise{Pareto comparison over 12 terrain tiles. Each point averages five decomposition resolutions; higher mIoU and lower RMSE are better. \gls{clear} occupies the favorable high-mIoU/low-RMSE region relative to Grid, Hexagonal, Quadtree, and SLIC.}}
    \label{fig:pareto-front}
\end{figure}

\subsection{Structural and Semantic Performance}
\label{subsec:structural_and_semantic_perf}
This subsection evaluates how effectively \gls{clear} preserves geometric structure and semantic composition across diverse terrains.
Abstraction quality is assessed using three complementary metrics that capture structural fidelity and semantic consistency:
the \textit{Root Mean Square Error (RMSE)} for elevation reconstruction accuracy,
the \textit{mean Intersection-over-Union (mIoU)} for semantic alignment,
and the \textit{repeatability ratio}, defined as the fraction of overlapping polygons with Intersection-over-Union (IoU)~$\ge$~0.5, for spatial consistency.

\textbf{Dataset.}
\revise{We construct a benchmark of 12 terrain tiles (\(100{\times}100\) pixels; each \(3\,\mathrm{km}\times3\,\mathrm{km}=9\,\mathrm{km}^2\)) sampled from Google Earth Engine.}
Each tile combines elevation data from the {NASADEM} dataset~\cite{nasadem2020} and land-cover labels from the {North American Land Change Monitoring System (NALCMS)}~\cite{nalcms2020}.
Terrain complexity is quantified by a combined score~$
\Psi = \alpha\,\text{JSD}_{\text{norm}}(p) + (1-\alpha)F,
\label{eq:combined_complexity}
$~%
where \(p\) is the empirical land-cover distribution, \(\text{JSD}_{\text{norm}}(p) = \sqrt{\text{JSD}(p \| u)}\) quantifies distributional diversity via Jensen-Shannon divergence from uniform \(u\), \(\alpha\!\in[0,1]\) balances distributional and spatial components, and
\[
F = \frac{1}{|\Omega|}\sum_{(i,j)\in\Omega}\frac{1}{k^2-1}\sum_{(u,v)\in\mathcal{N}_{ij}}\mathbb{1}[c_{i,j}\neq c_{u,v}]
\]
measures local heterogeneity over neighborhood~$\mathcal{N}_{ij}$.
We set \(\alpha=0.5\) and \(k=5\) for balanced sensitivity to class diversity and spatial variation.
\revise{\textbf{Decomposition benchmark.} Using~$\Psi$, the 12 tiles are stratified from homogeneous single-class surfaces ($\Psi{=}0.08$) to heterogeneous multi-class mixes ($\Psi{=}0.36$), spanning 13 NALCMS landcover classes, 5 dominant landcover types, 5 geographic biomes, and mean grades from 2--34\%.}
\revise{NALCMS was selected for its direct 30\,m co-registration with NASADEM and its 19-class hierarchy spanning ecologically distinct North American biomes. Non-North-American validation is discussed as future work in the Discussion.}

\textbf{Geometric and Semantic Fidelity.}
Elevation RMSE is computed between the original DEM and planar reconstructions from \autoref{alg:plane_fitting}, while mIoU measures agreement with ground-truth land-cover maps~\cite{cordts_cityscapes_2016,everingham_pascal_2015}.
Qualitative examples in \autoref{fig:elevation-reconstruction} and \autoref{fig:landcover-reconstruction} show contour-aligned, coherent regions that preserve topographic structure and fine land-cover transitions, unlike the blocky artifacts of Grid, Hexagonal, and Quadtree.

\revise{\autoref{fig:pareto-front} shows \gls{clear} on the favorable high-mIoU/low-RMSE side of the tradeoff relative to Grid, Hexagonal, Quadtree, and SLIC~\cite{achanta2012slic}. SLIC's spatial-color compactness objective does not enforce terrain-specific semantic/elevation alignment.}

\textbf{Repeatability.}
\revise{Across 10 trials on 90\%-overlapping Wharton patch pairs, BSD achieves \textbf{0.97~$\pm$~0.005} mean IoU and \textbf{96.8~$\pm$~0.6\%} repeatability, versus \textbf{0.39~$\pm$~0.01} and \textbf{4.5~$\pm$~3.8\%} for Quadtree, confirming stable boundary-seeded regions for map stitching and multi-robot planning.}

\begin{table}[t]
\centering
\scriptsize
\setlength{\tabcolsep}{2.4pt}
\caption{\revise{Planning results on W/H/R. Shaded rows (A*, AMRA*, RRT*): non-abstraction baselines; unshaded rows (Grid, Hex, Quadtree, CLEAR): abstraction methods. Bold/underline: best/second-best. Grade Fail: fraction of paths with $\geq$1 violation at 35\% grade and mean count. Region counts: W=31,440, H=156,004, R=218,074.}}

\label{tab:map_path_planning_3_maps}
\newcommand{\gc}{\cellcolor{gray!15}}
\resizebox{\columnwidth}{!}{%
\begin{tabular}{l l r r r r r r}
\toprule
Map & Method & Cost & Len (km) & P-Time(s) & A-Time(s) & Succ.(\%) & GF(\%, count) \\
\midrule

& \multicolumn{7}{l}{\textbf{\textit{No Abstraction}}} \\
 & \gc A*     & \gc 1.01 $\pm$ 0.07 & \gc 9.12 $\pm$ 0.72    & \gc 0.46 $\pm$ 0.24 & \gc ---           & \gc ---   & \gc 0.0, 0.00 \\
 & \gc AMRA*  & \gc 1.01 $\pm$ 0.07 & \gc 9.12 $\pm$ 0.72    & \gc 0.36 $\pm$ 0.02 & \gc ---           & \gc ---   & \gc 0.0, 0.00 \\
 W & \gc RRT*   & \gc 0.92 $\pm$ 0.16 & \gc 8.26 $\pm$ 0.97   & \gc 1.03 $\pm$ 0.02 & \gc ---           & \gc 85.00 & \gc 100.0, 9.31 $\pm$ 2.34 \\
 & \multicolumn{7}{l}{\textbf{\textit{Abstraction}}} \\
 & Grid     & 1.13 $\pm$ 0.07 & \underline{9.89} $\pm$ 0.67   & \textbf{0.16 $\pm$ 0.01} & \textbf{31.52} & --- & 20.0, 5.00 $\pm$ 0.00 \\
 & Hex      & \textbf{1.02 $\pm$ 0.07} & 8.71 $\pm$ 0.66          & \underline{0.22} $\pm$ 0.01 & 43.60    & --- & \textbf{0.0, 0.00} \\
 & Quadtree & 1.10 $\pm$ 0.07 & 8.49 $\pm$ 0.54                & \underline{0.22} $\pm$ 0.04 & 44.93    & --- & 100.0, 4.28 $\pm$ 2.50 \\
 & CLEAR    & \underline{1.05} $\pm$ 0.10 & \textbf{9.12 $\pm$ 1.15} & 0.23 $\pm$ 0.07 & \underline{33.02} & --- & \textbf{0.0, 0.00} \\
\midrule

& \multicolumn{7}{l}{\textbf{\textit{No Abstraction}}} \\
 & \gc A*     & \gc 4.11 $\pm$ 1.36 & \gc 37.50 $\pm$ 12.69 & \gc 8.06 $\pm$ 5.93  & \gc ---           & \gc ---   & \gc 0.0, 0.00 \\
 & \gc AMRA*  & \gc 4.11 $\pm$ 1.36 & \gc 37.50 $\pm$ 12.68 & \gc 5.00 $\pm$ 4.63  & \gc ---           & \gc ---   & \gc 0.0, 0.00 \\
 H & \gc RRT*   & \gc 4.25 $\pm$ 1.25 & \gc 36.06 $\pm$ 11.96 & \gc 10.07 $\pm$ 0.04 & \gc ---           & \gc 75.00 & \gc 70.0, 12.79 $\pm$ 2.94 \\
 & \multicolumn{7}{l}{\textbf{\textit{Abstraction}}} \\
 & Grid     & 4.78 $\pm$ 1.95 & 42.87 $\pm$ 18.35              & 0.91 $\pm$ 0.82 & 305.15          & --- & 0.0, 0.00 \\
 & Hex      & \underline{4.49} $\pm$ 1.57 & 38.55 $\pm$ 13.22  & 0.95 $\pm$ 0.75 & \underline{291.60} & --- & 0.0, 0.00 \\
 & Quadtree & 4.23 $\pm$ 1.44 & \underline{37.27} $\pm$ 12.47  & \underline{0.88} $\pm$ 0.42 & 486.61 & --- & 65.0, 3.50 $\pm$ 1.89 \\
 & CLEAR    & \textbf{4.41 $\pm$ 1.52} & \textbf{37.74 $\pm$ 12.67} & \textbf{0.87 $\pm$ 0.45} & \textbf{284.27} & --- & \textbf{0.0, 0.00} \\
\midrule

& \multicolumn{7}{l}{\textbf{\textit{No Abstraction}}} \\
 & \gc A*     & \gc 7.21 $\pm$ 1.43 & \gc 64.94 $\pm$ 12.70 & \gc 24.44 $\pm$ 11.52 & \gc ---          & \gc ---   & \gc 0.0, 0.00 \\
 & \gc AMRA*  & \gc 7.21 $\pm$ 1.43 & \gc 64.94 $\pm$ 12.70 & \gc 22.03 $\pm$ 13.77 & \gc ---          & \gc ---   & \gc 0.0, 0.00 \\
 R & \gc RRT*   & \gc 7.37 $\pm$ 1.49 & \gc 63.16 $\pm$ 13.26 & \gc 30.16 $\pm$ 0.07  & \gc ---          & \gc 90.00 & \gc 90.0, 11.81 $\pm$ 1.77 \\
 & \multicolumn{7}{l}{\textbf{\textit{Abstraction}}} \\
 & Grid     & 9.08 $\pm$ 2.21 & 80.18 $\pm$ 19.84              & 1.46 $\pm$ 1.29 & 631.44          & --- & 0.0, 0.00 \\
 & Hex      & \underline{8.15} $\pm$ 1.69 & 68.72 $\pm$ 13.32  & 1.56 $\pm$ 1.65 & \textbf{615.20} & --- & 0.0, 0.00 \\
 & Quadtree & 7.46 $\pm$ 1.50 & \underline{64.50} $\pm$ 12.91  & \underline{1.24} $\pm$ 0.34 & 834.59 & --- & 90.0, 4.74 $\pm$ 1.91 \\
 & CLEAR    & \textbf{7.85 $\pm$ 1.64} & \textbf{65.81 $\pm$ 13.39} & \textbf{1.25 $\pm$ 0.34} & \underline{652.26} & --- & \textbf{0.0, 0.00} \\
\bottomrule
\end{tabular}
}
\end{table}

\begin{figure*}[!t]
    \centering
    \scalebox{1}[1.0]{%
    \begin{tikzpicture}[every node/.style={inner sep=0, outer sep=0}]
        \node[anchor=south west] (imgA) at (0,0)
            {\includegraphics[width=0.73\textwidth]{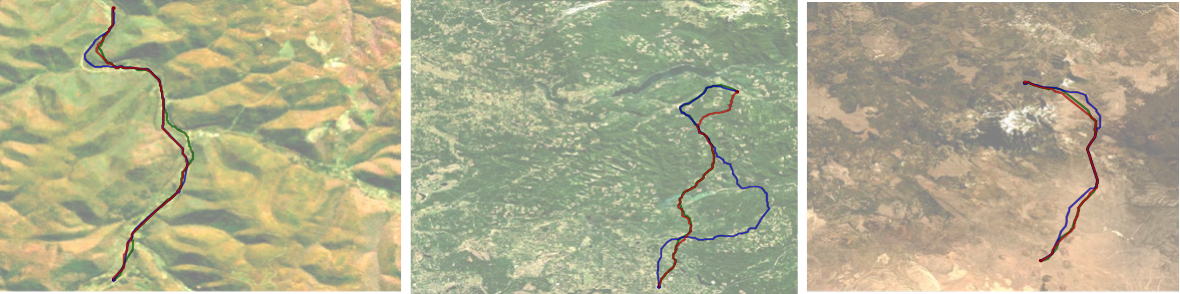}};
        \node[anchor=south west] (imgB) at ([xshift=0.74\textwidth]0,0)
            {\includegraphics[width=0.27\textwidth]{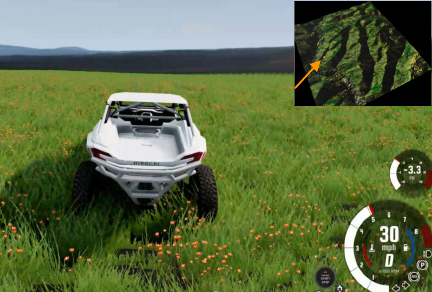}};
        \node[anchor=south west] at ([xshift=22mm,yshift=-5mm]imgA.south west) {\textbf{(a)}};
        \node[anchor=south west] at ([xshift=65mm,yshift=-5mm]imgA.south west) {\textbf{(b)}};
        \node[anchor=south west] at ([xshift=110mm,yshift=-5mm]imgA.south west) {\textbf{(c)}};
        \node[anchor=south west] at ([xshift=22mm,yshift=-5mm]imgB.south west) {\textbf{(d)}};
    \end{tikzpicture}
    }
    \vspace{-6mm}
    \caption{\revise{Planning and simulation overview. Paths on (a)  \textbf{W} (9\,km$^2$), (b) \textbf{H} (50\,km$^2$), and (c) \textbf{R} (100\,km$^2$) for \textcolor[RGB]{0,100,0}{\gls{clear}}, \textcolor{blue}{Quadtree}, and \textcolor{red}{dense-grid A*}. (d) BeamNG setup on W using NASADEM elevation and rasterized landcover.}}
    \label{fig:paths_and_sim}
\end{figure*}

\begin{figure}[!t]
\centering
\includegraphics[width=\linewidth]{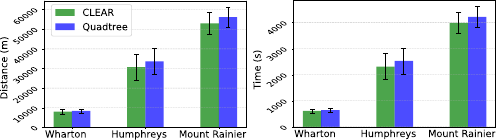}
\vspace{-6mm}
\caption{\revise{BeamNG executability on \textbf{W}, \textbf{H}, and \textbf{R}: distance and execution time for start--goal pairs where both planners complete. \gls{clear} produces shorter, faster runs across all maps. Summary in \autoref{tab:sim_results}.}}
\label{fig:sim_clear_quadtree_comparison}
\end{figure}

\begin{table}[!t]
\centering
\scriptsize
\setlength{\tabcolsep}{8pt}
\renewcommand{\arraystretch}{1.0}
\setlength{\aboverulesep}{0.9pt}
\setlength{\belowrulesep}{0.9pt}
\caption{\revise{BeamNG closed-loop executability on W/H/R. \gls{clear} achieves 100\% completion across all maps while Quadtree fails on W (90.2\%) and R (94.5\%).}}
\label{tab:sim_results}
\vspace{-1mm}
\begin{tabular}{l l r r l}
\toprule
\textbf{Map} & \textbf{Planner} & \textbf{Distance (m)} & \textbf{Time (s)} & \textbf{Completion} \\
\midrule
\multirow{2}{*}{W} & Quadtree  & 8376.16 & 645.23 & 90.2\% \\
& CLEAR  & \textbf{7977.01} & \textbf{609.70} & \textbf{100.0\%} \\
\addlinespace[1pt]
\cmidrule(lr){1-5}
\multirow{2}{*}{H}  & Quadtree  & 33574.32 & 2528.89 & 100.0\% \\
& CLEAR  & \textbf{30687.74} & \textbf{2313.76} & \textbf{100.0\%} \\
\addlinespace[1pt]
\cmidrule(lr){1-5}
\multirow{2}{*}{R}  & Quadtree  & 56132.00 & 4224.20 & 94.5\% \\
& CLEAR  & \textbf{53006.25} & \textbf{3994.53} & \textbf{100.0\%} \\
\bottomrule
\end{tabular}
\end{table}

\subsection{Path Planning and Executability}
\label{subsec:path_planning_executability}

\textbf{Planning Efficiency.}
Planning performance is evaluated using two metrics: (i)~\textit{planning time} for computational efficiency and (ii)~\textit{path cost} for feasibility and optimality.

\revise{\textbf{Planning maps.} Experiments use \textbf{Wharton (W)}, \textbf{Humphreys (H)}, and \textbf{Mount Rainier (R)}, spanning 9--100\,km$^2$ with diverse landcover and slope profiles: W is broadleaf/mixed forest with urban and cropland patches, H is high-elevation needleleaf/shrubland/grassland with urban patches, and R is needleleaf/shrubland/cropland with urban patches and broad relief.}

\revise{For each map, 20 diverse start--goal pairs are evaluated using the cost function defined in Sec.~\ref{sec:cost_function}.}
\revise{Grid, Hexagonal, and Quadtree are abstraction baselines evaluated under the same post-decomposition pipeline as \gls{clear}, so differences in results isolate the effect of spatial decomposition. A*, AMRA*~\cite{saxena2022amra}, and RRT* serve as non-abstraction planning references operating directly on the raw raster.}

\revise{Region budgets are matched across all abstraction baselines. For W/H/R, Quadtree recursively subdivides mixed-landcover cells until each leaf is landcover-pure or reaches a minimum area of 1/4/8 raster cells, scaled with map size to limit fragmentation. Let $N_Q$ denote the resulting leaf count. Grid uses spacing $s=\mathrm{round}(\sqrt{HW/N_Q})$, Hexagonal uses $s_h=\sqrt{2HW/(\sqrt{3}N_Q)}$ with row spacing $s_h\sqrt{3}/2$ and alternating half-row offsets, and \gls{clear} sets its BSD seed count to $n=N_Q$.}
\revise{RRT* uses the same vehicle cost model with a 35\% grade penalty. Its runtime budget is capped at the maximum planner runtime on each map plus a small buffer to permit path refinement without unbounded sampling. Remaining map-level parameters were selected through pilot sweeps and held fixed across all queries with no per-query retuning.}\footnote{For W/H/R, respectively, RRT* uses three attempts per query; time budgets of 1/10/30\,s; expansion radii of 5/15/25 pixels; rewiring radii of 15/45/75 pixels; goal-sampling rates of 20/20/25\%; and straight-line seeding.}

\begin{figure}[!t]
    \centering
    \includegraphics[width=\linewidth]{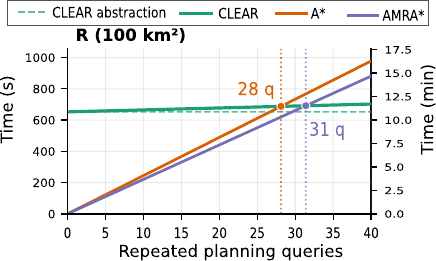}
    \vspace{-2mm}
    \caption{\revise{Rainier amortization over repeated planning queries. Markers indicate where \gls{clear}'s cumulative abstraction-plus-query time equals A* and AMRA*.}}
    \label{fig:amortization_rainier}
\end{figure}

\revise{\gls{clear} stays within 9\% of A*/AMRA* cost with zero slope violations on all maps, while Quadtree attains slightly lower nominal cost on H/R but violates grade constraints on 65\% and 90\% of paths. Per-query planning is reduced by $2.0\times$/$9.3\times$/$19.6\times$ over raw-grid A* on W/H/R, and abstraction cost amortizes after ${\sim}$144/${\sim}$40/${\sim}$28 queries versus A* and ${\sim}$254/${\sim}$69/${\sim}$31 versus AMRA* (\autoref{fig:amortization_rainier}).}

\begin{figure*}[!t]
    \centering
    \scalebox{1}[0.9]{%
    \begin{tikzpicture}
        \node[anchor=south west,inner sep=0] (image)
            {\includegraphics[width=\textwidth]{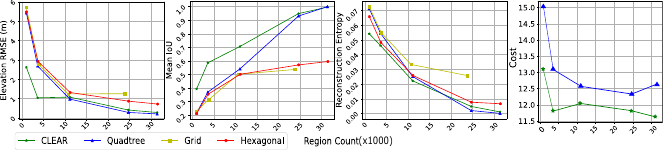}};
        \node[anchor=north west] at (2.0,4.5) {(a)};
        \node[anchor=north west] at (7.0,4.5) {(b)};
        \node[anchor=north west] at (11.5,4.5) {(c)};
        \node[anchor=north west] at (16.5,4.5) {(d)};
    \end{tikzpicture}
    }
    \vspace{-6mm}
    \caption{Ablation on \textbf{W}: (a--c) RMSE, mIoU, JSD vs.\ region count; (d) planning cost vs.\ region count.} %
    \label{fig:ablation-decomposition}
\end{figure*}

\begin{figure}[!t]
\vspace{-4mm}
    \centering
    \scalebox{1}[1.0]{%
    \includegraphics[width=\linewidth,]{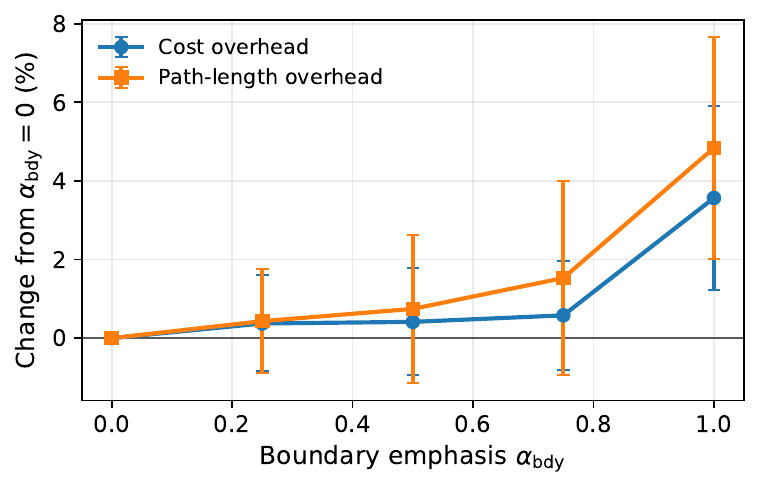}
    }
    \vspace{-4mm}
  \caption{\revise{Planning sensitivity to boundary emphasis ($\alpha_{\text{bdy}}$) on \textbf{H}, reported relative to $\alpha_{\text{bdy}}{=}0$. Cost changes by $<1\%$ through $\alpha_{\text{bdy}}{=}0.75$.}}
  \label{fig:ablate-boundary-emphasis}
\end{figure}

\textbf{Cost-agnostic evaluation.}
\revise{\gls{clear} remains close to the dense-raster A* lower bound under TERP's learned cost with comparable path lengths (\autoref{tab:terp_no_grade_table2}).}

\begin{table}[!t]
\centering
\scriptsize
\setlength{\tabcolsep}{6pt}
\renewcommand{\arraystretch}{1.0}
\setlength{\aboverulesep}{0.9pt}
\setlength{\belowrulesep}{0.9pt}
\caption{\revise{TERP-cost planning. Cost is scaled by $10^4$, and entries are mean $\pm$ standard deviation.}}
\label{tab:terp_no_grade_table2}
\vspace{-2mm}

\begin{tabular}{l l r r}
\toprule
Map & Method & TERP Cost & Path Len (m) \\
\midrule
\rowcolor{gray!15} \multirow{3}{*}{W} & A* & 0.88 $\pm$ 0.11 & 8132.66 $\pm$ 1156.60 \\
 & Quadtree & 0.97 $\pm$ 0.11 & 8645.46 $\pm$ 1193.96 \\
 & CLEAR & \textbf{0.92 $\pm$ 0.11} & \textbf{8235.54 $\pm$ 1135.99} \\
\midrule
\rowcolor{gray!15} \multirow{3}{*}{H} & A* & 3.71 $\pm$ 1.24 & 36552.89 $\pm$ 12041.32 \\
 & Quadtree & 3.80 $\pm$ 1.29 & 36532.29 $\pm$ 12079.41 \\
 & CLEAR & \textbf{3.80 $\pm$ 1.28} & \textbf{36204.80 $\pm$ 12012.91} \\
\midrule
\rowcolor{gray!15} \multirow{3}{*}{R} & A* & 6.26 $\pm$ 1.24 & 61431.53 $\pm$ 12067.04 \\
 & Quadtree & 6.47 $\pm$ 1.37 & 61925.21 $\pm$ 12765.85 \\
 & CLEAR & \textbf{6.45 $\pm$ 1.28} & \textbf{60990.40 $\pm$ 11822.67} \\
\bottomrule
\end{tabular}
\end{table}

\textbf{Onboard deployability.}
\revise{On a Jetson AGX Orin (JetPack~6.0, Ubuntu~22.04, 12-core Cortex-A78AE CPU), we time only graph search over precomputed abstractions (\autoref{tab:jetson_planning_time}). \gls{clear} is faster than Quadtree on W/H and only marginally slower on R (6.13 vs.\ 5.25\,s), where it avoids Quadtree's 90\% grade-failure rate.}

\begin{table}[!t]
\centering
\scriptsize
\setlength{\tabcolsep}{6pt}
\renewcommand{\arraystretch}{1.0}
\setlength{\aboverulesep}{0.9pt}
\setlength{\belowrulesep}{0.9pt}
\caption{\revise{Jetson AGX Orin query time. Planning time is mean $\pm$ standard deviation in seconds.}}
\label{tab:jetson_planning_time}
\begin{tabular}{l l r}
\toprule
Map & Decomp. & Planning Time (s) \\
\midrule
\rowcolor{gray!15} \multirow{3}{*}{W} & A* & 1.16 $\pm$ 0.25 \\
 & Quadtree & 0.77 $\pm$ 0.23 \\
 & CLEAR & \textbf{0.61} $\pm$ 0.15 \\
\midrule
\rowcolor{gray!15} \multirow{3}{*}{H} & A* & 31.07 $\pm$ 26.14 \\
 & Quadtree & 3.46 $\pm$ 1.57 \\
 & CLEAR & \textbf{2.99} $\pm$ 1.59 \\
\midrule
\rowcolor{gray!15} \multirow{3}{*}{R} & A* & 100.38 $\pm$ 53.39 \\
 & Quadtree & \textbf{5.25} $\pm$ 1.51 \\
 & CLEAR & 6.13 $\pm$ 2.27 \\
\bottomrule
\end{tabular}
\vspace{-2mm}
\end{table}

\textbf{Executability in Simulation.}
\revise{Planned trajectories are executed in \emph{BeamNG.tech}~\cite{beamng_tech}, a soft-body physics simulator, on NASADEM/landcover digital twins (\autoref{fig:paths_and_sim}(b)). A PID speed--heading controller drives each path to a $2\,\mathrm{m}$/$15^\circ$ goal tolerance. Because 9--100\,km$^2$ field validation requires dedicated vehicles and logistics, BeamNG provides a reproducible substitute capturing slope, friction, and terrain dynamics.}

\revise{\gls{clear} shortens executed path length by \textbf{6.3\%} on average (max \textbf{8.6\%} on H) and reduces execution time by \textbf{6.5\%} versus Quadtree, achieving \textbf{100\%} completion across all maps. The largest reliability gap appears on R, where Quadtree's planning-time grade violations propagate directly to execution failures (\autoref{tab:sim_results}).}

\subsection{Ablation Studies}
\label{subsec:alpha_ablation}
We ablate (i)~the \textit{region budget} on~\textbf{W} (Fig.~\ref{fig:ablation-decomposition}) and (ii)~the \textit{boundary emphasis}~$\alpha_{bdy}$ on~\textbf{H} (Fig.~\ref{fig:ablate-boundary-emphasis}).

\textbf{Region Budget – Decomposition.}
\autoref{fig:ablation-decomposition}(a-c) sweeps region count (standardized via Quadtree thresholds \{16,8,4,2,1\}). \gls{clear} achieves the lowest RMSE, highest mIoU, and lowest JSD with fewer regions, sustaining geometric and semantic integrity under coarse abstractions.

\textbf{Region Budget – Planning.}
\autoref{fig:ablation-decomposition}(d) shows \gls{clear} yields lower path costs across the full region-count range, maintaining an advantage even at coarse budgets.

\textbf{Boundary Emphasis \(\alpha_{bdy}\).}
On H, higher $\alpha_{bdy}$ improves boundary alignment but slightly increases path cost, motivating $\alpha_{bdy}=0$ for planning and $\alpha_{bdy}=1$ for maximal reconstruction fidelity.

%% file: sec/conclusion.tex
\section{Discussion}
\label{sec:discussion}

\textbf{Limitations.} CLEAR assumes obstacle-free terrain where rigid clutter is masked as non-traversable, though agent-specific clearance models could relax this conservatively. Mixed-landcover regions use a dominant label that may smooth minority classes, an effect captured quantitatively in the mIoU and JSD metrics. Computational cost scales with point density, and the boundary emphasis parameter $\alpha_{bdy}$ requires task-dependent tuning rather than a single universal setting. On the largest map scale, per-query planning time on embedded hardware marginally exceeds that of Quadtree, though this overhead is offset by CLEAR's zero grade violations. Validation is limited to North American terrain using NASADEM and NALCMS, whose 30~m co-registration enables consistent elevation-semantic pairing unavailable in global datasets.

\textbf{Convexity and Optimality.} BSD guarantees convex regions via Voronoi tessellation and recursive axis-aligned splitting. CLEAR trades a measured 6.7\% cost overhead for substantially faster repeated-query planning. The convex region structure is directly compatible with Graphs of Convex Sets optimization~\cite{marcucci2024shortest}, which could tighten this gap. Field deployment on terrain-capable robots, broader geographic validation, and integration of richer learned traversability models are natural extensions.

\section{Conclusion}
\label{sec:conclusion}
\revise{We introduced \gls{clear}, a reusable terrain abstraction that converts semantic boundaries and elevation geometry into repeatable, convex region graphs for large-scale off-road planning. \gls{clear} improves semantic and geometric fidelity over all decomposition baselines while preserving repeatable regions for map reuse. The same abstraction delivers \(2.0{\times}\)--\(19.6{\times}\) faster repeated-query graph search than raw-grid A* at 6.7\% cost overhead, and in closed-loop simulation produces 5--8.6\% shorter executed paths with 100\% task completion across the evaluated scenarios. Together, these results support \gls{clear} as a planning-aware abstraction layer rather than only a compact map representation.}

\section*{Acknowledgment}
The authors thank BeamNG.tech for providing an academic license for the simulator used in this research.